\documentclass[letterpaper, 10 pt, conference]{ieeeconf}  

\IEEEoverridecommandlockouts                              

\usepackage{caption}
\usepackage{times}
\usepackage{dsfont}
\usepackage{epsfig}
\usepackage{graphicx}
\usepackage{amsmath}
\usepackage{amssymb}
\usepackage{verbatim}
\usepackage{mathtools}
\let\labelindent\relax
\usepackage{enumitem}
\usepackage{booktabs}
\usepackage{xcolor}
\usepackage{multirow}
\usepackage{subcaption}
\usepackage{algorithm} 
\usepackage{algpseudocode} 
\usepackage{adjustbox}
\usepackage[detect-all]{siunitx}
\usepackage[pagebackref=true,breaklinks=true,letterpaper=true,colorlinks,citecolor=blue,linkcolor=blue,bookmarks=false]{hyperref}

\usepackage{pifont}

\title{\LARGE \bf
Improving Generalization Ability for 3D Object Detection by Learning Sparsity-invariant Features
}

\author{Hsin-Cheng Lu$^{1}$, Chung-Yi Lin$^{1}$ and Winston H. Hsu$^{1, 2}$
\thanks{$^{1.}$ National Taiwan University
}
\thanks{$^{2.}$ Mobile Drive Technology}
}

\begin{document}

\maketitle
\thispagestyle{empty}
\pagestyle{empty}

\begin{abstract}

In autonomous driving, 3D object detection is essential for accurately identifying and tracking objects. Despite the continuous development of various technologies for this task, a significant drawback is observed in most of them—they experience substantial performance degradation when detecting objects in unseen domains. In this paper, we propose a method to improve the generalization ability for 3D object detection on a single domain. We primarily focus on generalizing from a single source domain to target domains with distinct sensor configurations and scene distributions. To learn sparsity-invariant features from a single source domain, we selectively subsample the source data to a specific beam, using confidence scores determined by the current detector to identify the density that holds utmost importance for the detector. Subsequently, we employ the teacher-student framework to align the Bird's Eye View (BEV) features for different point clouds densities. We also utilize feature content alignment (FCA) and graph-based embedding relationship alignment (GERA) to instruct the detector to be domain-agnostic. Extensive experiments demonstrate that our method exhibits superior generalization capabilities compared to other baselines. Furthermore, our approach even outperforms certain domain adaptation methods that can access to the target domain data. The code is available at \url{https://github.com/Tiffamy/3DOD-LSF}.

\end{abstract}

\section{INTRODUCTION}

\label{sec:intro}

The perception stage is crucial for numerous real-world applications, including robotics navigation and autonomous driving. As LiDAR sensors can provide precise depth information compared to RGB cameras, they have gained increasing popularity in the field of perception. Among those perception tasks, 3D object detection aims to detect and localize objects of interest in the surrounding environment. With the advent of 3D point cloud deep learning technologies~\cite{qi2017pointnet, qi2017pointnet++} and the release of several 3D real-world human-annotated datasets~\cite{geiger2012we, caesar2020nuscenes, sun2020scalability}, numerous research studies related to 3D object detection models~\cite{zhou2018voxelnet, yan2018second, yang2018pixor, lang2019pointpillars, shi2019pointrcnn, shi2020pv, shi2020points, shi2020point, yin2021center, deng2021voxel, shi2023pv} have emerged recently in the pursuit of achieving more accurate results. However, most of those 3D object detection works focus on training in a specific domain. They might experience a significant performance drop when used in an unknown domain because they may not account for different sensor settings and environments during the training phase, thereby limiting their applicability for real-world applications.

\begin{figure}[t!]
\centering
\includegraphics[width=1.0\columnwidth, clip]{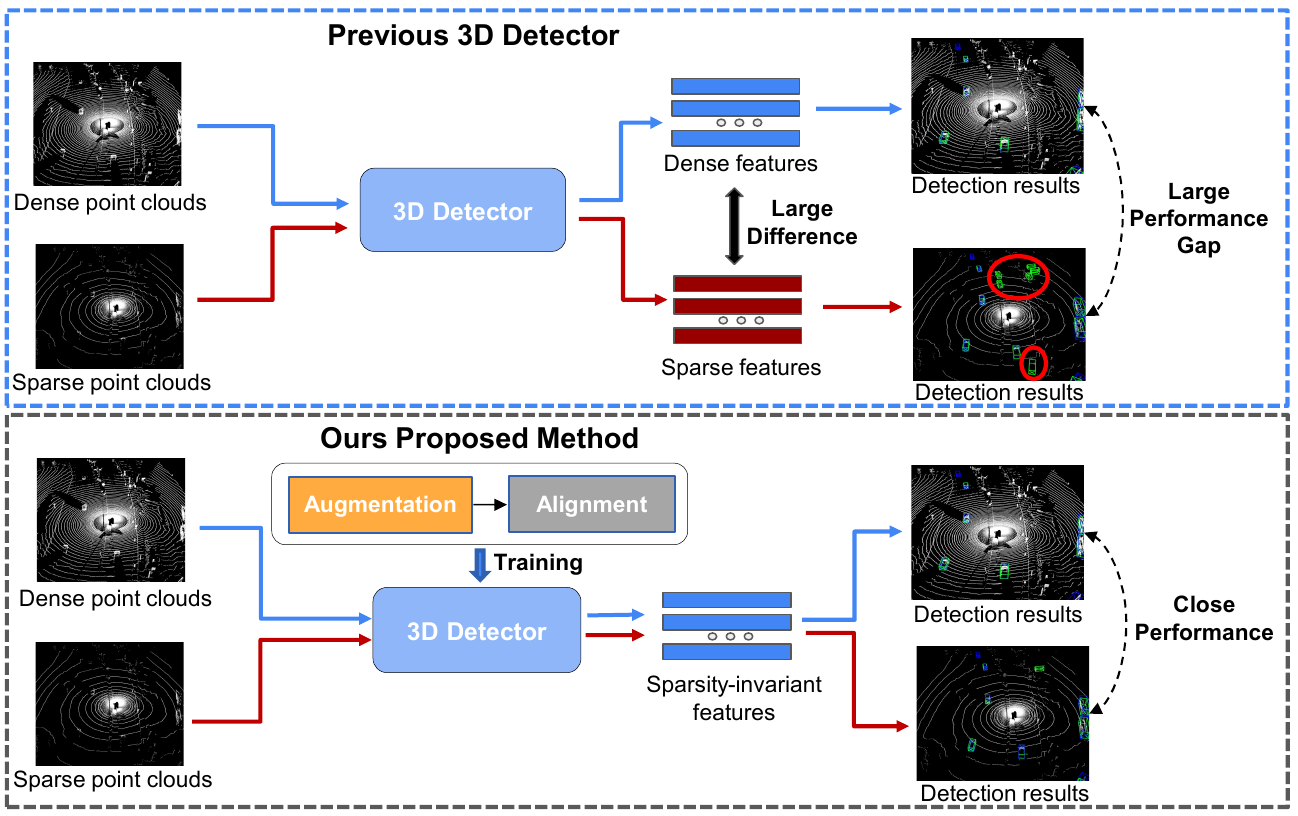}
\caption{Traditional 3D detectors, trained directly on the source domain, often experience a significant performance drop when the point clouds become sparser. In contrast, our method empowers the 3D detector to learn sparsity-invariant features through training with our proposed augmentation and feature alignment techniques. Note that in the detection results, blue boxes represent the ground truth annotations, while the green boxes indicate the predicted boxes.
}

\vspace{-.5cm}
\label{fig1}
\end{figure}

As mentioned in~\cite{wang2020train}, the domain gap in 3D can be categorized into two main reasons. The first reason is that the LiDAR sensor configurations for each dataset are different, leading to significant differences in the sensed point clouds. The other reason arises from scene distribution. Public datasets collect their data from different environments. All these factors contribute to varying statistical distributions, including the distribution of each class, the shape, and the size of each object. Table~\ref{dataset_overview} summarizes the collection details for each dataset.

To mitigate the domain gap between various 3D datasets, some works~\cite{wang2020train, saltori2020sf, yang2021st3d, xu2021spg, yihan2021learning, wei2022lidar, hu2023density, tsai2023viewer} have proposed unsupervised domain adaptation (UDA) methods for 3D object detection, which aims to transfer the detector trained on source domain to a new domain. Since these methods primarily focus on enhancing performance in the target domain, they cannot ensure high performance in unseen domains, rendering them less applicable for real-world usage. To this end, domain generalization (DG) aims to improve the generalization ability of 3D detectors, ensuring performance in unseen domains and robustness against domain gaps. Existing works~\cite{lehner20223d} use data augmentation to enhance model robustness against damaged or unusually shaped cars. However, DG techniques addressing other general domain issues, such as sensor configuration, remain relatively unexplored.

In this work, we propose a novel method to improve the generalization ability of a 3D object detector on a single source domain. As shown in Fig.~\ref{fig1}, the detector directly trained on the source domain experiences a significant performance drop when the point clouds density changes. Consequently, our work primarily focuses on mitigating the domain shift caused by different sensor configurations. Additionally, due to the efficiency and practicality of low-beam LiDAR for real-world applications, we generalize the detector from a high-beam source domain to low-beam target domains. We first selectively downsample the source data to a specific density, using confidence scores determined by the current detector to identify the density that holds utmost importance for the detector. With the augmented domain and the original domain, we employ the student-teacher framework with our proposed graph-based embedding relationship alignment (GERA) and feature content alignment (FCA) to instruct the model to overlook the density difference between the augmented domain and the original domain. GERA ensures that the high-level pairwise relationships remain consistent, while FCA serves to maintain low-level content consistency. Through training with these techniques, the 3D object detector becomes more robust to domain shifts.

To assess the generalization ability of the detector trained with our method, we conduct experiments across various datasets, including Waymo~\cite{sun2020scalability}, KITTI~\cite{geiger2012we}, and nuScenes~\cite{caesar2020nuscenes}. Experimental results demonstrate that the detector trained using our method exhibits superior generalization ability. Remarkably, even without access to any target domain data, our method shows comparable performance to UDA methods. Another experiment also indicates that model trained with our method can also collaborate with UDA methods to achieve better performance on the target domain.

Our contributions are summarized as follows:
\begin{itemize}
  \item We propose a method to improve generalization ability for 3D object detection on a single source domain.
  \item We introduce an augmentation strategy to downsample the point cloud to a specific sparsity based on the confidence of the detector.
  \item We develop graph-based embedding relationship alignment along with feature content alignment to maintain high-level relationship and low-level content consistency for domain-invariant representation learning.
  \item Extensive experiments show that our method outperforms all the baselines in terms of generalization. Furthermore, it can be compatible with UDA techniques to achieve even better performance on the target domain.
\end{itemize}

\begin{table}[t]
    \scriptsize
    \centering

    \caption{Datasets overview. The dataset size refers to the total number of annotated frames.}
    \resizebox{\columnwidth}{!}{\begin{tabular}{ c | c c c c c}
        \hline
        Dataset & Size & \begin{tabular}[x]{@{}c@{}}LiDAR\\beams\end{tabular} & \begin{tabular}[x]{@{}c@{}}Points Per\\Beam\end{tabular} & \begin{tabular}[x]{@{}c@{}}Vertival\\ FOV(°)\end{tabular} & Location \\
        \hline \hline
        Waymo~\cite{sun2020scalability} & 230K & 64 & 2258 & [-17.6°, 2.4°] & USA\\ 
        KITTI~\cite{geiger2012we} & 15K & 64 & 1863 & [-23.6°, 3.2°] & Germany\\  
        nuScenes~\cite{caesar2020nuscenes} & 40K & 32 & 1084 & [-30.0°, 10.0°] & USA/Singapore \\   
        \hline
    \end{tabular}}

    \label{dataset_overview}
\end{table}

\section{RELATED WORK}

\begin{figure*}[t!]
\centering
\includegraphics[width=1.0\textwidth]{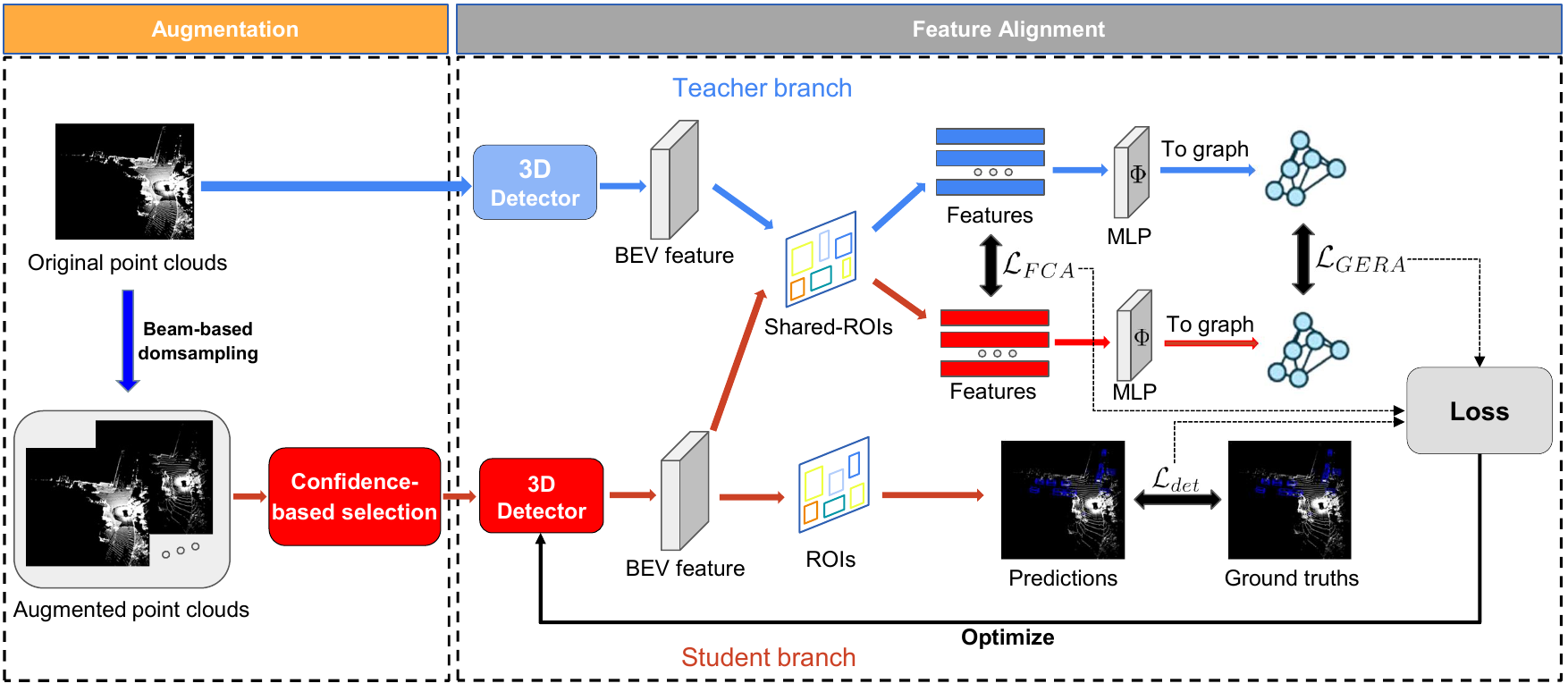}
\caption{ The overview of our proposed method. Initially, we downsample original point clouds into various densities and then select one based on detector confidence. To learn domain-agnostic features, feature content alignment (FCA) are applied to BEV features for each shared Region of Interest (ROI) to align low-level content consistency. Subsequently, encoded features are constrained by graph-based embedding relationship alignment (GERA) to maintain high-level relationship consistency. The blue and red flows illustrate the processing pipelines for the teacher and student models, respectively.}

\vspace{-10pt}
\label{fig2}
\end{figure*}

\label{relatedwork}

\subsection{LiDAR-based 3D Object Detection}
LiDAR-based 3D Object Detection aims to estimate precise positions and orientations of 3D objects in scene-level point clouds. In terms of representations used, the existing works can be categorized into voxel-based~\cite{zhou2018voxelnet, yan2018second, yang2018pixor, lang2019pointpillars, shi2020points, deng2021voxel, yin2021center}, point-based~\cite{shi2019pointrcnn, shi2020point} and point-voxel based~\cite{shi2020pv, shi2023pv} methods. VoxelNet~\cite{zhou2018voxelnet} rasterizes point clouds into volumetric dense grids, followed by 3D CNNs that convolve along each dimension. SECOND~\cite{yan2018second} leverages sparse convolution to eliminate unnecessary computation wasted on unoccupied zero-padding voxels. Point-RCNN~\cite{shi2019pointrcnn} is the first to introduce a method for generating 3D proposals from point clouds in a bottom-up fashion. PV-RCNN~\cite{shi2020pv} integrate the advantages of both point-based and voxel-based methods to produce more precise results. In our work, We choose SECOND~\cite{yan2018second} and PV-RCNN~\cite{shi2020pv} as our backbone detectors.

\subsection{Unsupervised Domain Adaptation for 3D object detection}
Unsupervised Domain Adaptation (UDA) aims to leverage both the labeled source domain and the unlabeled target domain to transfer knowledge from the source domain to the target domain. \cite{wang2020train} is a pioneer in investigating the domain gap between LiDAR datasets for 3D object detection. ST3D~\cite{yang2021st3d} employs the self-training strategy and trains networks with curriculum data augmentation. LiDAR Distillation~\cite{wei2022lidar} and DTS~\cite{hu2023density} utilize the teacher-student framework with augmentation strategies related to point cloud density resampling to address the domain gap caused by different LiDAR configurations. However, UDA techniques still aim to improve performance in a specific target domain, and they cannot guarantee performance in other unseen domains.

\subsection{Domain Generalization}
The goal of Domain Generalization (DG) is to improve the performance on unseen domains that are not used in the training process. In 2D computer vision, various DG works have been proposed for object detection~\cite{wu2022single, vidit2023clip}. Nevertheless, DG for 3D vision tasks is still an under-explored problem. Uni3D~\cite{zhang2023uni3d} strengthen feature reusability across different datasets to generalize the detector from multiple domains. Given the cost-expensiveness of obtaining multiple fully-labeled LiDAR datasets as source domains for training, we decide to improve the generalization ability only from a single source domain. Previous works~\cite{yang2021st3d, choi2021part, lehner20223d, hu2023density} have introduced their augmentation strategies to make detectors robust to domain shifts. ROS from \cite{yang2021st3d} randomly scales object sizes for diverse object detection. RBRS in \cite{hu2023density} augments point clouds via density re-sampling, emphasizing density distribution manipulation for improved model robustness. PA-AUG~\cite{choi2021part} is proposed to obtains the robustness to corrupted data for 3D detection models. 3D-VField~\cite{lehner20223d} adversarially augment each car to make the detector more robust to the rare shaped and damaged car. While their emphasis is on data augmentation, we combine our augmentation strategy and the student-teacher framework with proposed feature alignment techniques to achieve better generalization ability.

\section{PROBLEM FORMULATION}

Suppose we have a source domain $D^{s} = \{(X^{s}_{i}, Y^{s}_{i})\}_{i=1}^{N_{s}}$ and an unseen target domain $D^{t} = \{(X^{t}_{i}, Y^{t}_{i})\}_{i=1}^{N_{t}}$, where s and t represent source and target domain respectively. $N_{s}$ is the number of source domain samples while $X^{s}_{i}$ means the $i$-th source point cloud and $Y^{s}_{i}$ is it's corresponding label. The label $Y^{s}_{i}$ is parameterized by its center $(c_{x}, c_{y}, c_{z})$, size $(l, w, h)$, and the heading angle $\theta$. Our objective is to train a 3D object detector on the source domain $D^{s}$, enhancing its generalization ability so that it excels not only in the source domain but also demonstrates strong performance on unseen domain $D^{t}$.

\section{LEARNING SPARSITY-INVARIANT FEATURES FOR 3D OBJECT DETECTION}

\label{method}

\subsection{Point Clouds Downsampling} 
\begin{figure}[t!]
\centering
\includegraphics[width=1.0\columnwidth, clip]{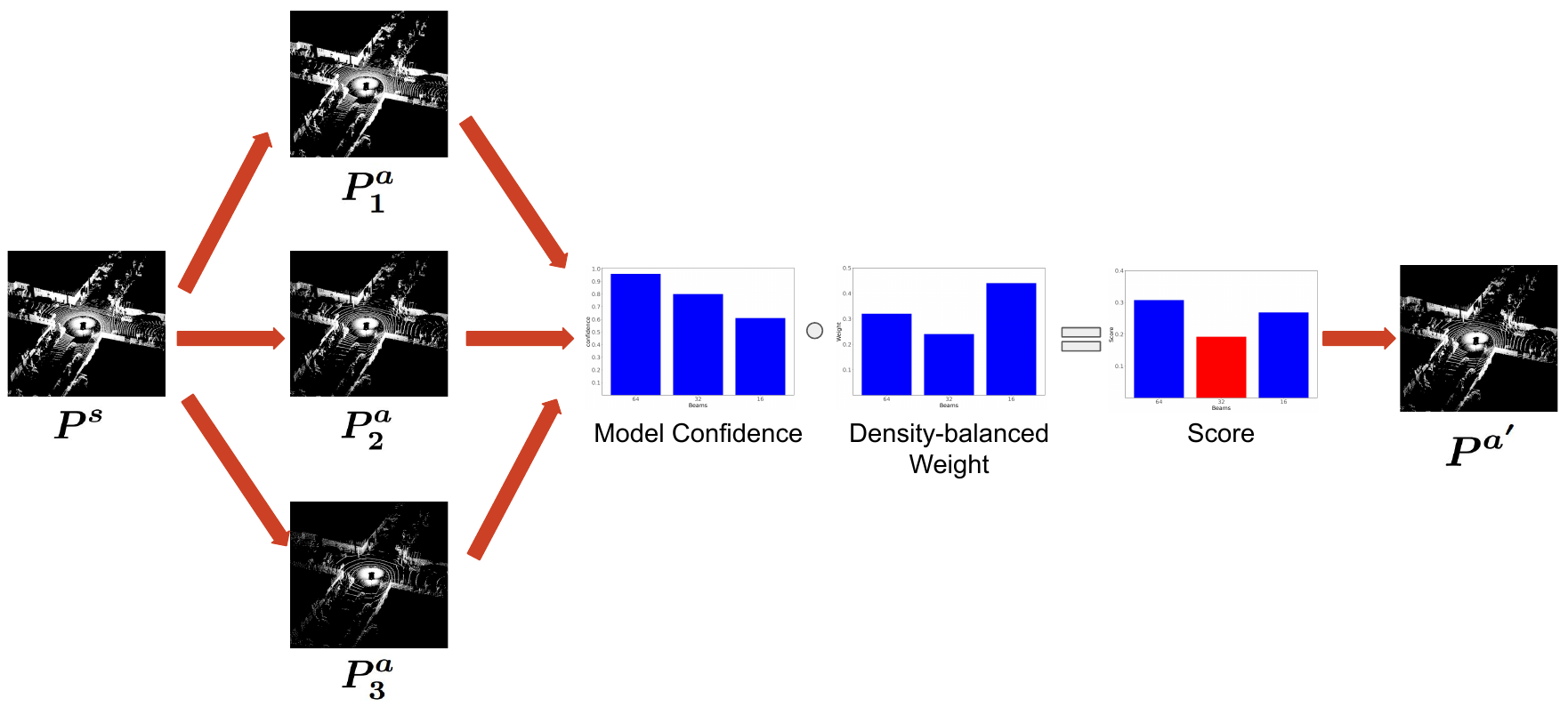}
\caption{Illustration of our confidence-based selection strategy. $P^{s}$ and $P^{a'}$ denote the original point clouds and the final augmented point clouds, respectively, while $\{P^{a}_{i}\}_{i=1}^{N}$ indicates $N$ augmented point clouds with different beam types.}

\vspace{-.5cm}
\label{fig3}
\end{figure}

Learning generalizable features from a single source domain is a challenging task, as we can only acquire knowledge within point clouds captured under same sensor configurations and similar scene distribution. To address this challenge, we can leverage our prior knowledge of LiDAR sensors to simulate point clouds across various densities.
\begin{enumerate}[label=\emph{\arabic*)}, wide]
    \item \emph{Beam-based Downsampling:} To downsample point clouds into specific beam types, we adopt the approach from \cite{wei2022lidar}. We first convert the Cartesian coordinates $(x, y, z)$ of points to spherical coordinates using the equations:
    \begin{equation}
        \begin{split}
        \theta = \arctan \frac{z}{\sqrt{x^2 + y^2}},\\
        \phi = \arctan \frac{y}{\sqrt{x^2 + y^2}},
        \end{split}
        \label{coord_trans}
    \end{equation}
    where $\theta$ and $\phi$ represent zenith and azimuth angles. For beam-based downsampling, it is necessary to split the points from different beams. Since some datasets lack beam labels for each LiDAR beam, the K-means algorithm is then applied on the zenith angle $\theta$ to assign beam labels to each point. With the obtained labels for each LiDAR beam, downsampling the original point clouds to specific types of point clouds corresponding to different beams becomes straightforward.
    
    \item \emph{Confidence-based Selection:} After downsampling original point clouds to point clouds with distinct densities, we then require a strategy to select one of them. We observe that randomly choosing one may result in an imbalanced performance across various densities. This is because random selection may sometimes lead to choosing a density that the model is already familiar with, diverting the model from the intended goal of generalization. To address this issue, we propose confidence-based selection to choose the density that that holds utmost importance for the current detector. Given the original point clouds $P^{s}$ and sets of augmented point clouds $\{P^{a}_{i}\}_{i=1}^{N}$, where $N$ is the number of augmented point clouds with different beam types, we can inference our current detector on  $\{P^{a}_{i}\}_{i=1}^{N}$ to obtain the matched objects $\{O_{i, j}\}_{j=1}^{N_{a}}$ for the $i$-th augmented point cloud by computing the Intersection over Union (IoU) between each predicted object and the ground truth object, where $N_{a}$ is the number of matched objects. Note that the matching process associates each predicted object with a ground truth object having the maximum IoU. With matched objects $\{O_{i, j}\}_{j=1}^{N_{a}}$, we can calculate the confidence for each augmented data using the following equation:
    \begin{equation}
        S{i} = \frac{\sum_{j=1}^{N_{a}}{\mathds{1}_{IoU>IoU_{th}} C_{i,j}}}{\sum_{j=1}^{N_{a}}{\mathds{1}_{IoU>IoU_{th}}}},
    \end{equation}
    where $C_{i,j}$ represents the confidence for the $j$-th matched object in the $i$-th augmented point cloud, and $S_{i}$ is the final confidence score for the $i$-th augmented point cloud. The term $IoU$ denotes the matched IoU for the object, and $IoU_{th}$ indicates the threshold for considering a matched object in the computation. To achieve a balance between the quantity and confidence for various densities, we additionally weighted the confidence score $S_{i}$ with the proportion of data we have already selected for each beam type. Subsequently, we select the augmented point cloud with the lowest score as our augmented domain. Fig.~\ref{fig3} illustrate the pipeline for our confidence-based selection strategy.
\end{enumerate}

With the augmented domain and the original domain, we employ the student-teacher framework for sparsity-invariant features learning. To ensure the accuracy of our teacher model on the original domain and guide the student model to generalize across various density levels, we feed the original data to the teacher model and the augmented data to the student model. Notice that the initial teacher model and the student model are pretrained on the original source domain with our proposed augmentation.

\subsection{Feature Content Alignment}
The purpose of employing the student-teacher framework is to train the detector to learn domain-invariant features. Since the only difference between the original domain and the augmented domain is the LiDAR beam number, the features generated by the teacher model fed with the original data and the student model fed with the augmented data should be similar. To this end, we can directly align the feature content between the student detector and the teacher detector.

Given the BEV features generated by both the student and teacher models, we use the shared-ROI map to extract instance proposal features from both sets of BEV features. This results in student instance proposal features $F^{a} \in \mathbb{R}^{N_{r} \times H \times W \times M}$ and teacher instance proposal features $F^{s} \in \mathbb{R}^{N_{r} \times H \times W \times M}$, where $N_{r}$ is the number of region proposals. $H$ and $W$ represent the height and width for a region proposal, and $M$ is the feature dimension. We then easily align these instance proposal features by the following equation:
\begin{equation}
    \mathcal{L}_{FCA} = \frac{\sum_{i=1}^{N_{r}}{\|f^{s}_{i} - f^{a}_{i}\|_{2}}}{N_{r}},
\end{equation}
where $f^{s}_{i}$ and $f^{a}_{i}$ are the $i$-th feature of $F^{s}$ and $F^{a}$, respectively.

\subsection{Graph-based Embedding Relationship Alignment}
By employing feature content alignment, we enforce the model to generate similar features regardless of the density change. However, it is also crucial for the relationship between each instance proposal feature pair to remain consistent, given that we only alter the beam type of the original point clouds.

To measure the discrepancy of the pairwise relationships, we can transform the features for each proposal into a fully-connected graph. This allows us to focus on computing the edge difference between two graphs constructed by the student instance proposal features and the teacher instance proposal features, respectively.

To construct a graph from the instance proposal features, we first convert the three-dimensional features $f^{s}_{i}$ and $f^{a}_{i}$ into one-dimensional embeddings by passing them through a multi-perception layer (MLP) to capture the high-level meanings for each proposal. Here, $f^{s}_{i}$ and $f^{a}_{i}$ represent the $i$-th feature of $F^{s}$ and $F^{a}$ mentioned above, respectively. The graph is then built, with the edges representing the cosine similarity between embeddings. The process can be described as:
\begin{equation}
    E^{s}_{i, j} = \frac{\Phi(f^{s}_{i}) \cdot \Phi(f^{s}_{j})}{\|\Phi(f^{s}_{i})\| \|\Phi(f^{s}_{j})\|},
\end{equation}
where $E^{s}_{i, j}$ denotes the edge between $i$-th embedding and $j$-th embedding for teacher and $\Phi$ is a MLP utilized to extract high-level embeddings from the three-dimensional features.

Given the edges $E^{s}_{i, j}$ and $E^{a}_{i, j}$ for teacher and student, respectively, we then utilize Gromov-Wasserstein discrepancy~\cite{peyre2016gromov} to define our graph-based embedding relationship alignment loss as the following equation:
\begin{equation}
    \mathcal{L}_{GERA} = \sum_{i, j, m, n} KL(E^{a}_{i,j}, E^{s}_{m, n})R_{i, m}R_{j, n},
\end{equation}
where $KL$ indicates the Kullback-Leibler divergence to measure the distance of the edges across graphs. $R$ is the relationship matrix to represent the difference distance between each proposal.

In addition to the completely matched edge pairs, the relationship between edge pairs containing similar instance proposals should also remain consistent across graphs. To address this, we initially construct a discrepancy matrix by calculating the differences in center, size, and orientation between each instance proposal. The discrepancy matrix $D$ is then formulated as below:
\begin{equation}
    D_{i,m} = \frac{1}{\|c_{i}-c_{m}\|_{2}+\|d_{i}-d_{m}\|_{2}+\|r_{i}-r_{m}\|_{1}+\epsilon},
\end{equation}
where $c_{i}$, $d_{i}$ and $r_{i}$ denote the center, size, and orientation for the $i$-th proposal, respectively. $\epsilon$ is a factor used to scale the discrepancy value and prevent zero division errors.

After computing the discrepancy matrix, we define the relationship matrix $R$ as:
\begin{equation}
    R = I + \lambda D,
\end{equation}
where $I$ is the identity matrix, preserving the weights for the totally matched edge pairs, and  $\lambda$ controls the importance of the discrepancy matrix.

\subsection{Overall Loss Function}
In the student-teacher framework, we freeze the parameters of the teacher model, while the parameters of the student model are updated by the original detection loss $\mathcal{L}_{det}$, feature content alignment loss $\mathcal{L}_{FCA}$, and graph-based embedding relationship alignment loss $\mathcal{L}_{GERA}$. This can be expressed as the following equation:
\begin{equation}
\mathcal{L}_{overall} = \mathcal{L}_{det} + \alpha \mathcal{L}_{FCA} + \beta \mathcal{L}_{GERA},
\end{equation}
Where $\alpha$ and $\beta$ are hyperparameters to balance two alignment losses. 

With these two alignments, our goal is to guide the student detector to align both the low-level content and the high-level relationship between the augmented domain and the original domain for sparsity-invariant feature learning.

\section{EXPERIMENTS}

\label{experiment}

\begin{table*}[t!]
    \small
    \centering
        \caption{Generalization results from the Waymo dataset using two different detection backbones. The reported AP is the moderate case for KITTI and the overall result for other datasets. The best score across all baselines is underlined, while the best score excluding domain adaptation (DA) methods is indicated in \textbf{bold}. }

        \begin{adjustbox}{max width=\textwidth}
        \begin{tabular}
            { l | c | c c | c | c | c c | c }
            \toprule[1.5pt]
             \multirow{2}{*}{Methods} & 
             \multicolumn{4}{c|}{SECOND-IoU}  & 
             \multicolumn{4}{c}{PV-RCNN}
             \\
             
             \cline{2-9}  &
             W$^{1}$ & $\rightarrow$N$^{2}$ & $\rightarrow$K$^{3}$ & Avg. & 
             W & $\rightarrow$N & $\rightarrow$K & Avg.   
             \\ \hline
    
             Source Only & \textbf{\underline{67.75}/\underline{55.68}} & 32.91/17.24 & 67.64/27.48 & 56.10/33.47
             & \textbf{\underline{70.11}/\underline{58.56}} & 34.50/21.47 & 61.18/22.01 & 55.26/34.11 \\ \hline
             
             ROS~\cite{yang2021st3d} & 65.36/51.60 & 30.86/17.28 & 77.93/51.40 & 58.05/40.09
             & 62.24/47.23 & 29.79/18.85 &  78.82/\textbf{56.01} & 56.95/40.70 \\

             PA-AUG~\cite{choi2021part} & 67.50/53.98 & 28.79/17.47 & 77.23/46.12 & 57.84/39.19 
             & 68.01/56.76 & 32.11/20.06 & 71.91/38.39 & 57.34/38.40 \\
             
             RBRS~\cite{hu2023density} & 67.25/52.13 & 38.82/21.76 & 76.41/43.32 & 60.83/39.07
             & 68.08/56.10 & 41.58/25.12 & 77.31/34.72 & 62.32/38.65 \\ 
             
             Ours & 66.58/51.27 & \textbf{39.45/22.30} & \textbf{79.96/53.60} & \textbf{\underline{62.00}/\underline{42.39}}
             & 67.68/56.03 & \textbf{41.88/25.83} & \textbf{80.03}/48.25 & \textbf{\underline{63.20}/\underline{43.37}} \\ \hline
             
            ST3D($\rightarrow$N)~\cite{yang2021st3d} & 61.94/46.80 & 35.92/20.19 & 69.57/39.14 & 55.81/35.38 
            & 65.53/54.04 & 36.42/22.99 & 70.49/41.19 & 57.48/39.41 \\
            
            ST3D($\rightarrow$K)~\cite{yang2021st3d} & 53.22/34.13 & 25.54/10.19 & 82.19/61.83 & 53.65/35.38 
            & 59.37/42.34 & 30.59/18.44 & 84.10/64.78 & 58.02/41.85 \\
    
            DTS($\rightarrow$N)~\cite{hu2023density} & 42.67/30.56 & \underline{41.20}/\underline{23.00} & 66.45/30.32 & 50.11/27.96 
            & 46.10/39.99 & \underline{44.00}/\underline{26.20} & 77.69/39.27 & 55.93/35.15 \\
    
            DTS($\rightarrow$K)~\cite{hu2023density} & 49.18/31.38 & 27.43/18.48 & \underline{85.80}/\underline{71.50} & 54.14/40.45
            & 54.37/38.27 & 27.52/18.37 & \underline{86.40}/\underline{68.10} & 56.10/41.58 \\
            \bottomrule[1.5pt]
        \end{tabular}
        \end{adjustbox}
        \\
$^{1}$ denotes Waymo, $^{2}$ denotes transfer or adapt to nuScenes, and $^{3}$ denotes transfer or adapt to KITTI.
    \vspace{-.2cm}

    \label{table:MainResult}
\end{table*}

\subsection{Experiment Details}
\textbf{Datasets.} We conduct our experiments on three well-known autonomous driving datasets: Waymo~\cite{sun2020scalability}, KITTI~\cite{geiger2012we}, and nuScenes~\cite{caesar2020nuscenes}. As shown in Table~\ref{dataset_overview}, Waymo is the most dense and largest dataset. 
We choose to generalize mainly from the Waymo dataset for most of our experiments, while also utilizing the KITTI dataset to demonstrate generalization results by only altering the LiDAR beams for the original point clouds.


\textbf{Metrics.} Following previous UDA work~\cite{yang2021st3d}, we adopt the KITTI evaluation metric for assessing our methods on the commonly used car category in KITTI, nuScenes, and the vehicle category in Waymo. We report the average precision (AP) over 40 recall positions under the IoU threshold of 0.7 for both the bird’s eye view (BEV) IoUs and 3D IoUs. To demonstrate the generalization ability, we also compute the average of AP over all datasets. We also report \textbf{Closed Gap}~\cite{yang2021st3d} to demonstrate how much our method can help close the performance gap between Source Only and Oracle under UDA settings, which is defined as: $\textbf{Closed Gap} = \frac{AP_{model} - AP_{source~only}}{AP_{oracle} - AP_{source~only}} \times 100\% $. Note that all experiments evaluate the methods on the validation set.

\subsection{Comparison with State-of-the-art Methods}
We benchmark our method against both DG and UDA baselines as shown in Table~\ref{table:MainResult}. For DG baselines, we compare our method with augmentation strategies designed to enhance the detector's robustness to domain shifts, including ROS \textbf{(augmentation strategy of ST3D~\cite{yang2021st3d})}, RBRS \textbf{(augmentation strategy of DTS~\cite{hu2023density})}, and PA-AUG~\cite{choi2021part}. For DA baselines, we compare our method with ST3D~\cite{yang2021st3d} and DTS~\cite{hu2023density}.

We observe that previous augmentation techniques achieve better performance compared to Source only. ROS excels in the Waymo$\rightarrow$KITTI setting, addressing the primary domain gap related to size distribution. PA-AUG exhibits limited performance improvement, primarily benefiting objects with rare shapes. RBRS enhances the performance in the Waymo$\rightarrow$nuScenes scenario due to significant differences in point cloud density. Our method incorporates our proposed augmentation  and alignments strategies, which facilitates the learning of domain-invariant features, resulting in superior performance across unseen domains.

When comparing with UDA baselines, it is evident that while they perform well in their target domains, their performance across other domains is not guaranteed. In the Waymo$\rightarrow$nuScenes task, our method demonstrates comparable performance to UDA methods and even outperforms ST3D, which employs self-training for target domain data. This highlights the significance of improving the generalization ability of the detector. A robust generalization ability is also crucial for adapting to new domains, even when access to them is available.

We also validate our method with different detection backbones. This indicated that our method compatible with both the lightweight detector SECOND-IoU~\cite{yang2021st3d} and the two-stage detector PV-RCNN~\cite{shi2020pv}, which provides better precision.

\subsection{Ablation Study}
\begin{table}[t!]
    \centering
    \caption{Ablation study for our method. BDS refers to beam-based downsampling and CBS is the confidence-based selection strategy. The best score is \textbf{bolded}.}
    \begin{tabular}{c c | c c | c | c c | c}
    \toprule[1.5pt]

    \multicolumn{2}{c|}{Augmentation} & \multicolumn{2}{c|}{Alignments}
    & \multicolumn{4}{c}{$\SI{}{AP_{3D}}$} \\ \hline

    BDS & CBS & FCA & GERA & W$^{1}$ & $\rightarrow$N$^{2}$ & $\rightarrow$K$^{3}$ & Avg. \\ \hline

    & & & & \textbf{55.68} & 17.24 & 27.48 & 33.47 \\

    $\checkmark$ & & & & 52.87 & 20.15 & 43.92 & 38.98 \\
    $\checkmark$ & $\checkmark$ & & & 51.97 & 20.83 & 46.82 & 39.87 \\ \hline
    $\checkmark$ & $\checkmark$ & $\checkmark$ & & 50.90 & 21.50 & 51.96 & 41.45 \\
    $\checkmark$ & $\checkmark$ & & $\checkmark$ & 53.36 & 21.04 & 48.65 & 41.02 \\
    $\checkmark$ & $\checkmark$ & $\checkmark$ & $\checkmark$ & 51.27 & \textbf{22.30} & \textbf{53.60} & \textbf{42.39} \\
    
    \bottomrule[1.5pt]
    \end{tabular}

$^{1}$ denotes Waymo , $^{2}$ denotes transfer to nuScenes, and $^{3}$ denotes transfer to KITTI.
    \label{tab:ablation}
\end{table}

    

To validate the effectiveness of each component for our method, we conduct the ablation study experiment as shown in Table~\ref{tab:ablation}. All experiments are conducted with the detector SECOND-IoU and use Waymo as the source domain.  

In the augmentation phase, applying only the beam-based downsampling significantly enhances overall generalization ability. However, this approach leads to unstable performance in unseen domains since the density used for training is randomly selected. By employing downsampling with the confidence selection strategy, the current detector dynamically chooses the density it needs to learn, leading to improved generalization ability by sacrificing a little performance on the original source domain. 

In our two alignments, FCA enforces the model to generate similar BEV features regardless of point cloud density, while GERA preserves high-level relationship consistency for instance pairs. Training with only FCA yields significant improvement for unseen domains; however, it struggles to preserve the performance on the original domain. On the other hand, the GERA loss enables the detector to learn how to maintain consistency in the relationships of instance embeddings, preserving more performance on the original domain but showing limited improvement in unseen domains.

Combining both alignment losses, our method achieves a balance between low-level and high-level consistency, resulting in the best generalization ability across all domains.

\begin{figure}[h!]
\centering
\includegraphics[width=1\columnwidth, clip]{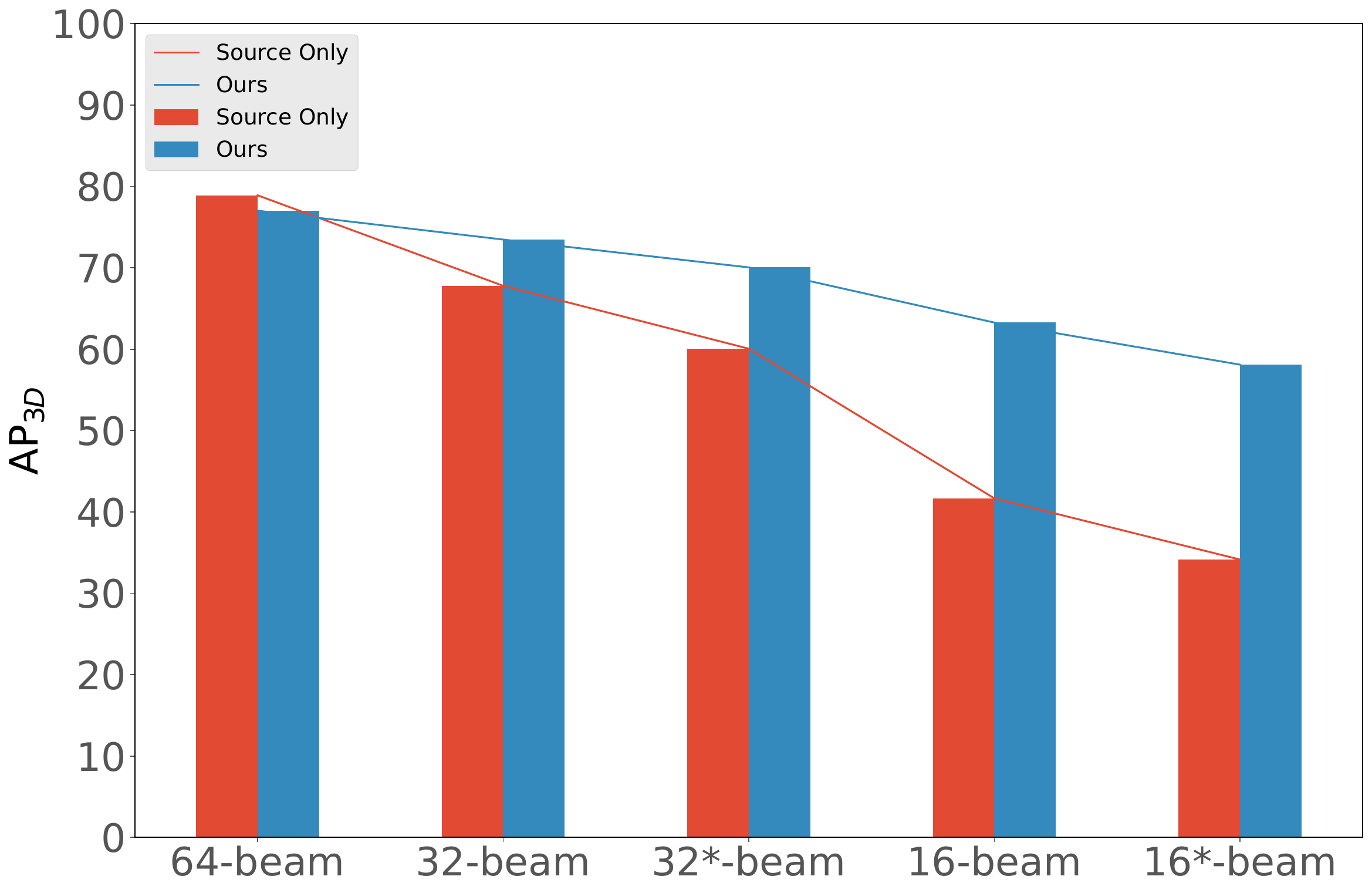}
\caption{Performance on KITTI dataset~\cite{geiger2012we} across various point cloud densities. Source Only indicates the direct evaluation of the model trained on the original dataset (64-beam) on other low-beam validation sets.  Ours denotes that the detector is trained using the proposed method. 
}
\label{fig4}
\end{figure}

\subsection{Single Dataset Generalization}
To validate the generalization ability of our method across different densities, we only modify the beam type of the original point clouds in this experiment. Note that we not only decrease the number of LiDAR beams but also reduce the number of points for each beam, \emph{i.e.}, 32*-beam means we subsample half of the points of the 32-beam data. In this experiment, we use SECOND-IoU as our backbone model. As shown in Fig.~\ref{fig4}, Source Only model suffers from a severe performance drop as the point cloud density becomes sparser, while our method maintains acceptable performance even when transferring to 16*-beam data. This experiment highlights that the model trained with our method becomes more robust to changes in point cloud density.

\subsection{Compatible with Domain Adaptation Methods}
\begin{table}[t!]
    \centering
    \caption{Domain adaptation of Waymo$\rightarrow$nuScenes setting. The best score is indicated in \textbf{bold}.}
    \begin{tabular}{l | c | c}
    \toprule[1.5pt]
         Methods & $\SI{}{AP_{BEV}}$/$\SI{}{AP_{3D}}$ & Closed Gap \\ \hline
         Source only & 32.91/17.24 & - \\
         Ours & 39.45/22.30 & +34.48\%/+28.70\% \\ \hline
         ST3D & 35.92/20.19 & +15.87\%/+16.73\% \\
         Ours(w/ ST3D) & \textbf{40.63/24.04} & \textbf{+40.70\%/+38.57\%} \\ \hline
         Oracle & 51.88/34.87 & - \\

    \bottomrule[1.5pt]
    \end{tabular}
    \label{tab:With_DA}
\end{table}

Since our method doesn't use any target domain data during training, we can further compatible our method with DA techniques. We conduct this experiment by utilizing SECOND-IoU as our backbone model and evaluate the performance in Waymo$\rightarrow$nuScenes adaptation setting. In this experiment, we also report \textbf{Closed Gap} as mentioned in experiment details. As shown in Table~\ref{tab:With_DA}, our method significantly boosts ST3D~\cite{yang2021st3d} by 4.71/3.85 and 24.83\%/21.84\% for $\SI{}{AP_{BEV}}$/$\SI{}{AP_{3D}}$ and Closed Gap, respectively. This underscores the importance of our approach even when target domain data is accessible.

\section{CONCLUSIONS}

\label{conclusion}

In this paper, we propose a method to enhance the generalization ability of LiDAR-based 3D object detection, making the detector more robust to changes in point clouds density. We introduce confidence-based downsampling to simulate point clouds under various densities based on the confidence score. We also adopt the student-teacher framework with our proposed alignment losses to preserve both the low-level content consistency and the high-level relationship consistency. Extensive experiments demonstrate that our method exhibits better generalization ability compared to other baselines.

\section{ACKNOWLEDGEMENT}

This work was supported in part by the National Science and Technology Council, Taiwan, under Grant NSTC 113-2634-F-002-007. We are grateful to the National Center for High-performance Computing.








\clearpage
\bibliographystyle{IEEEtran}
\bibliography{egbib}

\begin{thebibliography}{10}
\providecommand{\url}[1]{#1}
\csname url@rmstyle\endcsname
\providecommand{\newblock}{\relax}
\providecommand{\bibinfo}[2]{#2}
\providecommand\BIBentrySTDinterwordspacing{\spaceskip=0pt\relax}
\providecommand\BIBentryALTinterwordstretchfactor{4}
\providecommand\BIBentryALTinterwordspacing{\spaceskip=\fontdimen2\font plus
\BIBentryALTinterwordstretchfactor\fontdimen3\font minus \fontdimen4\font\relax}
\providecommand\BIBforeignlanguage[2]{{%
\expandafter\ifx\csname l@#1\endcsname\relax
\typeout{** WARNING: IEEEtran.bst: No hyphenation pattern has been}%
\typeout{** loaded for the language `#1'. Using the pattern for}%
\typeout{** the default language instead.}%
\else
\language=\csname l@#1\endcsname
\fi
#2}}

\bibitem{qi2017pointnet}
C.~R. Qi, H.~Su, K.~Mo, and L.~J. Guibas, ``Pointnet: Deep learning on point sets for 3d classification and segmentation,'' in \emph{Proceedings of the IEEE conference on computer vision and pattern recognition}, 2017, pp. 652--660.

\bibitem{qi2017pointnet++}
C.~R. Qi, L.~Yi, H.~Su, and L.~J. Guibas, ``Pointnet++: Deep hierarchical feature learning on point sets in a metric space,'' \emph{Advances in neural information processing systems}, vol.~30, 2017.

\bibitem{geiger2012we}
A.~Geiger, P.~Lenz, and R.~Urtasun, ``Are we ready for autonomous driving? the kitti vision benchmark suite,'' in \emph{2012 IEEE conference on computer vision and pattern recognition}.\hskip 1em plus 0.5em minus 0.4em\relax IEEE, 2012, pp. 3354--3361.

\bibitem{caesar2020nuscenes}
H.~Caesar, V.~Bankiti, A.~H. Lang, S.~Vora, V.~E. Liong, Q.~Xu, A.~Krishnan, Y.~Pan, G.~Baldan, and O.~Beijbom, ``nuscenes: A multimodal dataset for autonomous driving,'' in \emph{Proceedings of the IEEE/CVF conference on computer vision and pattern recognition}, 2020, pp. 11\,621--11\,631.

\bibitem{sun2020scalability}
P.~Sun, H.~Kretzschmar, X.~Dotiwalla, A.~Chouard, V.~Patnaik, P.~Tsui, J.~Guo, Y.~Zhou, Y.~Chai, B.~Caine, \emph{et~al.}, ``Scalability in perception for autonomous driving: Waymo open dataset,'' in \emph{Proceedings of the IEEE/CVF conference on computer vision and pattern recognition}, 2020, pp. 2446--2454.

\bibitem{zhou2018voxelnet}
Y.~Zhou and O.~Tuzel, ``Voxelnet: End-to-end learning for point cloud based 3d object detection,'' in \emph{Proceedings of the IEEE conference on computer vision and pattern recognition}, 2018, pp. 4490--4499.

\bibitem{yan2018second}
Y.~Yan, Y.~Mao, and B.~Li, ``Second: Sparsely embedded convolutional detection,'' \emph{Sensors}, vol.~18, no.~10, p. 3337, 2018.

\bibitem{yang2018pixor}
B.~Yang, W.~Luo, and R.~Urtasun, ``Pixor: Real-time 3d object detection from point clouds,'' in \emph{Proceedings of the IEEE conference on Computer Vision and Pattern Recognition}, 2018, pp. 7652--7660.

\bibitem{lang2019pointpillars}
A.~H. Lang, S.~Vora, H.~Caesar, L.~Zhou, J.~Yang, and O.~Beijbom, ``Pointpillars: Fast encoders for object detection from point clouds,'' in \emph{Proceedings of the IEEE/CVF conference on computer vision and pattern recognition}, 2019, pp. 12\,697--12\,705.

\bibitem{shi2019pointrcnn}
S.~Shi, X.~Wang, and H.~Li, ``Pointrcnn: 3d object proposal generation and detection from point cloud,'' in \emph{Proceedings of the IEEE/CVF conference on computer vision and pattern recognition}, 2019, pp. 770--779.

\bibitem{shi2020pv}
S.~Shi, C.~Guo, L.~Jiang, Z.~Wang, J.~Shi, X.~Wang, and H.~Li, ``Pv-rcnn: Point-voxel feature set abstraction for 3d object detection,'' in \emph{Proceedings of the IEEE/CVF conference on computer vision and pattern recognition}, 2020, pp. 10\,529--10\,538.

\bibitem{shi2020points}
S.~Shi, Z.~Wang, J.~Shi, X.~Wang, and H.~Li, ``From points to parts: 3d object detection from point cloud with part-aware and part-aggregation network,'' \emph{IEEE transactions on pattern analysis and machine intelligence}, vol.~43, no.~8, pp. 2647--2664, 2020.

\bibitem{shi2020point}
W.~Shi and R.~Rajkumar, ``Point-gnn: Graph neural network for 3d object detection in a point cloud,'' in \emph{Proceedings of the IEEE/CVF conference on computer vision and pattern recognition}, 2020, pp. 1711--1719.

\bibitem{yin2021center}
T.~Yin, X.~Zhou, and P.~Krahenbuhl, ``Center-based 3d object detection and tracking,'' in \emph{Proceedings of the IEEE/CVF conference on computer vision and pattern recognition}, 2021, pp. 11\,784--11\,793.

\bibitem{deng2021voxel}
J.~Deng, S.~Shi, P.~Li, W.~Zhou, Y.~Zhang, and H.~Li, ``Voxel r-cnn: Towards high performance voxel-based 3d object detection,'' in \emph{Proceedings of the AAAI Conference on Artificial Intelligence}, vol.~35, no.~2, 2021, pp. 1201--1209.

\bibitem{shi2023pv}
S.~Shi, L.~Jiang, J.~Deng, Z.~Wang, C.~Guo, J.~Shi, X.~Wang, and H.~Li, ``Pv-rcnn++: Point-voxel feature set abstraction with local vector representation for 3d object detection,'' \emph{International Journal of Computer Vision}, vol. 131, no.~2, pp. 531--551, 2023.

\bibitem{wang2020train}
Y.~Wang, X.~Chen, Y.~You, L.~E. Li, B.~Hariharan, M.~Campbell, K.~Q. Weinberger, and W.-L. Chao, ``Train in germany, test in the usa: Making 3d object detectors generalize,'' in \emph{Proceedings of the IEEE/CVF Conference on Computer Vision and Pattern Recognition}, 2020, pp. 11\,713--11\,723.

\bibitem{saltori2020sf}
C.~Saltori, S.~Lathuili{\'e}re, N.~Sebe, E.~Ricci, and F.~Galasso, ``Sf-uda 3d: Source-free unsupervised domain adaptation for lidar-based 3d object detection,'' in \emph{2020 International Conference on 3D Vision (3DV)}.\hskip 1em plus 0.5em minus 0.4em\relax IEEE, 2020, pp. 771--780.

\bibitem{yang2021st3d}
J.~Yang, S.~Shi, Z.~Wang, H.~Li, and X.~Qi, ``St3d: Self-training for unsupervised domain adaptation on 3d object detection,'' in \emph{Proceedings of the IEEE/CVF conference on computer vision and pattern recognition}, 2021, pp. 10\,368--10\,378.

\bibitem{xu2021spg}
Q.~Xu, Y.~Zhou, W.~Wang, C.~R. Qi, and D.~Anguelov, ``Spg: Unsupervised domain adaptation for 3d object detection via semantic point generation,'' in \emph{Proceedings of the IEEE/CVF International Conference on Computer Vision}, 2021, pp. 15\,446--15\,456.

\bibitem{yihan2021learning}
Z.~Yihan, C.~Wang, Y.~Wang, H.~Xu, C.~Ye, Z.~Yang, and C.~Ma, ``Learning transferable features for point cloud detection via 3d contrastive co-training,'' \emph{Advances in Neural Information Processing Systems}, vol.~34, pp. 21\,493--21\,504, 2021.

\bibitem{wei2022lidar}
Y.~Wei, Z.~Wei, Y.~Rao, J.~Li, J.~Zhou, and J.~Lu, ``Lidar distillation: Bridging the beam-induced domain gap for 3d object detection,'' in \emph{European Conference on Computer Vision}.\hskip 1em plus 0.5em minus 0.4em\relax Springer, 2022, pp. 179--195.

\bibitem{hu2023density}
Q.~Hu, D.~Liu, and W.~Hu, ``Density-insensitive unsupervised domain adaption on 3d object detection,'' in \emph{Proceedings of the IEEE/CVF Conference on Computer Vision and Pattern Recognition}, 2023, pp. 17\,556--17\,566.

\bibitem{tsai2023viewer}
D.~Tsai, J.~S. Berrio, M.~Shan, E.~Nebot, and S.~Worrall, ``Viewer-centred surface completion for unsupervised domain adaptation in 3d object detection,'' in \emph{2023 IEEE International Conference on Robotics and Automation (ICRA)}.\hskip 1em plus 0.5em minus 0.4em\relax IEEE, 2023, pp. 9346--9353.

\bibitem{lehner20223d}
A.~Lehner, S.~Gasperini, A.~Marcos-Ramiro, M.~Schmidt, M.-A.~N. Mahani, N.~Navab, B.~Busam, and F.~Tombari, ``3d-vfield: Adversarial augmentation of point clouds for domain generalization in 3d object detection,'' in \emph{Proceedings of the IEEE/CVF conference on computer vision and pattern recognition}, 2022, pp. 17\,295--17\,304.

\bibitem{wu2022single}
A.~Wu and C.~Deng, ``Single-domain generalized object detection in urban scene via cyclic-disentangled self-distillation,'' in \emph{Proceedings of the IEEE/CVF Conference on computer vision and pattern recognition}, 2022, pp. 847--856.

\bibitem{vidit2023clip}
V.~Vidit, M.~Engilberge, and M.~Salzmann, ``Clip the gap: A single domain generalization approach for object detection,'' in \emph{Proceedings of the IEEE/CVF Conference on Computer Vision and Pattern Recognition}, 2023, pp. 3219--3229.

\bibitem{zhang2023uni3d}
B.~Zhang, J.~Yuan, B.~Shi, T.~Chen, Y.~Li, and Y.~Qiao, ``Uni3d: A unified baseline for multi-dataset 3d object detection,'' in \emph{Proceedings of the IEEE/CVF Conference on Computer Vision and Pattern Recognition}, 2023, pp. 9253--9262.

\bibitem{choi2021part}
J.~Choi, Y.~Song, and N.~Kwak, ``Part-aware data augmentation for 3d object detection in point cloud,'' in \emph{2021 IEEE/RSJ International Conference on Intelligent Robots and Systems (IROS)}.\hskip 1em plus 0.5em minus 0.4em\relax IEEE, 2021, pp. 3391--3397.

\bibitem{peyre2016gromov}
G.~Peyr{\'e}, M.~Cuturi, and J.~Solomon, ``Gromov-wasserstein averaging of kernel and distance matrices,'' in \emph{International conference on machine learning}.\hskip 1em plus 0.5em minus 0.4em\relax PMLR, 2016, pp. 2664--2672.

\end{thebibliography}

\end{document}